\documentclass[lettersize,journal]{IEEEtran}
\usepackage{amsmath,amsfonts}
\usepackage{algorithmic}
\usepackage{algorithm}
\usepackage{array}
\usepackage[caption=false,font=normalsize,labelfont=sf,textfont=sf]{subfig}
\usepackage{textcomp}
\usepackage{stfloats}
\usepackage{url}
\usepackage{verbatim}
\usepackage{graphicx}
\usepackage{cite}
\usepackage{multirow}                 
\usepackage{multicol}                 
\usepackage{multirow}                
\usepackage{float}                    
\usepackage{makecell}                 
\usepackage{booktabs}                 
\usepackage{amssymb}
\usepackage{amsthm}
\usepackage{xcolor}
\begin{document}

\title{M$^3$amba: CLIP-driven Mamba Model for Multi-modal Remote Sensing Classification}

\author{Mingxiang Cao,
Weiying Xie,~\IEEEmembership{Senior Member,~IEEE},	
Xin Zhang, 
Jiaqing Zhang,
Kai Jiang,
Jie Lei,~\IEEEmembership{Member,~IEEE},
Yunsong Li,~\IEEEmembership{Member,~IEEE}
\thanks{This work was supported in part by the National Natural Science Foundation of China under Grant 62121001, Grant 62322117 and Grant 62371365, 62071360, in part by the National Key R\&D Program of China (No. 2023YFE0208100), and supported by the Innovation Fund of Xidian University under Grant YJSJ25007. (Corresponding~author: Weiying Xie.)

Mingxiang Cao, Weiying Xie, Xin Zhang, Jiaqing Zhang, Kai Jiang, and Yunsong Li are with the State Key
Laboratory of Integrated Services Networks, Xidian University, Xi'an 710071,
China (e-mail: mingxiangcao@stu.xidian.edu.cn; wyxie@xidian.edu.cn; xinzhang\_xd@163.com; jqzhang\_2@stu.xidian.edu.cn; xdjiangkai@foxmail.com;  ysli@mail.xidian.edu.cn). 

Jie Lei is at the School of Electrical and Data Engineering at the University of Technology Sydney (e-mail: jie.lei@uts.edu.au).}}


\markboth{IEEE Transactions on Circuits and Systems for Video Technology}%
{Shell \MakeLowercase{\textit{et al.}}: A Sample Article Using IEEEtran.cls for IEEE Journals}


\IEEEpubid{\begin{minipage}{\textwidth}\ \centering
		Copyright \copyright 2025 IEEE. Personal use of this material is permitted. \\
		However, permission to use this material for any other purposes must be obtained 
		from the IEEE by sending an email to pubs-permissions@ieee.org.
\end{minipage}}

\maketitle

\begin{abstract}
Multi-modal fusion holds great promise for integrating information from different modalities. However, due to a lack of consideration for modal consistency, existing multi-modal fusion methods in the field of remote sensing still face challenges of incomplete semantic information and low computational efficiency in their fusion designs. Inspired by the observation that the visual language pre-training model CLIP can effectively extract strong semantic information from visual features, we propose M$^3$amba, a novel end-to-end CLIP-driven Mamba model for multi-modal fusion to address these challenges. Specifically, we introduce CLIP-driven modality-specific adapters in the fusion architecture to avoid the bias of understanding specific domains caused by direct inference, making the original CLIP encoder modality-specific perception. This unified framework enables minimal training to achieve a comprehensive semantic understanding of different modalities, thereby guiding cross-modal feature fusion. To further enhance the consistent association between modality mappings, a multi-modal Mamba fusion architecture with linear complexity and a cross-attention module Cross-SS2D are designed, which fully considers effective and efficient information interaction to achieve complete fusion. Extensive experiments have shown that M$^3$amba has an average performance improvement of at least 5.98\% compared with the state-of-the-art methods in multi-modal hyperspectral image classification tasks in the remote sensing field, while also demonstrating excellent training efficiency, achieving a double improvement in accuracy and efficiency. The code is released at https://github.com/kaka-Cao/M3amba.
\end{abstract}

\begin{IEEEkeywords}
Deep learning, Remote Sensing, Multi-modal, Feature Fusion, CLIP model, Mamba.
\end{IEEEkeywords}

\section{Introduction}
\begin{figure}[htbp]
	\centering
	\includegraphics[width=1\linewidth]{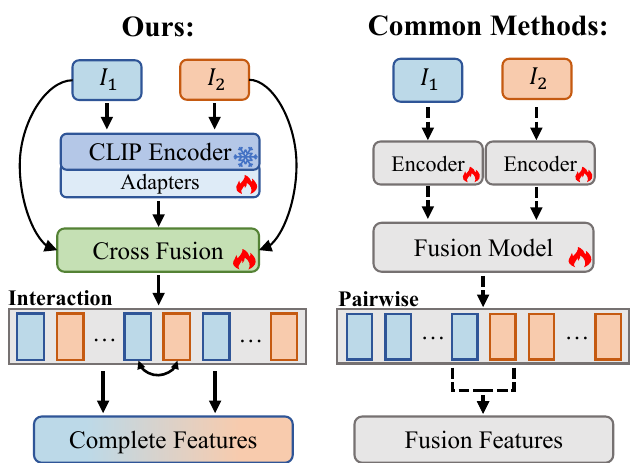}
	\caption{\textbf{Left:} Proposed CLIP-driven modality-specific adapters guide the fusion process to produce complete features through semantic information interactions. \textbf{Right:} Common methods perform pairwise fusion by training encoders and fusion networks, which lacks consideration of semantic consistency and leads to incomplete representation. \(I_1\) and \(I_2\) represent inputs from different modalities.}
	\label{intro_fig}
		\vspace{-10pt}
\end{figure}
Multi-modal fusion tasks involve integrating features from various data modalities, such as visible light, infrared, and LiDAR, to achieve more complete feature representations \cite{zhao2023tufusion, li2024feddiff, ma2023mbsi}. This is crucial for applications like remote sensing scene understanding \cite{li2024mdfl}. Unfortunately, due to hardware limitations, a single sensor cannot capture all the complex details of an image \cite{xie2024fusionmamba}, but the semantic information in multi-modal images is always consistent, so it is particularly important to effectively leverage comprehensive semantic information across different modalities to guide fusion \cite{zhang2024simfusion}. Existing methods focus on training different models for semantic feature extraction, but these models have limitations. CNN-based approaches cannot capture comprehensive semantic information due to their limited receptive fields, while Transformer-based methods have global modeling capabilities but incur high training costs and lack transferability. The advent of Contrastive Language-Image Pre-training (CLIP) \cite{radford2021learning} offers a solution to these issues. CLIP achieves significant success in various visual tasks by pre-training with a large corpus of image-text pairs obtained from the web. Given its robustness and transferability, CLIP is often used to directly infer effective image features. For example, Lin \textit{et al.} \cite{lin2022frozen} used frozen CLIP output features to capture semantic information between adjacent frames. Yang \textit{et al.} \cite{yang2024inf} used CLIP image embedding for global understanding, allowing the model to learn from high-level semantic information. This confirms CLIP's powerful ability to learn rich semantic visual representations and its ability to transfer to other visual tasks. However, direct inference often results in weak perception capabilities of downstream tasks \cite{gao2024clip}, and the potential of this approach in the field of multi-modal images remains underexplored.
\IEEEpubidadjcol

It is noteworthy that multi-modal image data captured and processed by different sensors often exhibit characteristics of consistency and complementarity \cite{zhang2020deep}. These traits can significantly enhance the capabilities of deep learning models, facilitating more detailed and nuanced task interpretation. Thus, obtaining complete multi-modal fusion features is crucial for optimizing tasks. Compared to traditional linear concatenation methods in CNN architectures, approaches based on attention mechanisms have shown satisfactory results. Ma \textit{et al.} \cite{ma2022swinfusion} introduced a novel universal image fusion framework based on cross-domain distance learning and Swin Transformer, achieving comprehensive integration and global interaction of multi-modal complementary information. Although the emergence of self-attention mechanisms has facilitated the interaction of information flow between modal features, they bring challenges of quadratic computational requirements and lack the rich semantic representation capabilities of multi-modal fusion features.
Thus, a natural question arises: Can we obtain comprehensive visual semantic information from different modalities with minimal computational overhead to guide a fusion network for complete feature fusion? To overcome the shortcomings of existing methods, we propose the first CLIP-driven \textbf{M}amba model for \textbf{M}ulti-\textbf{M}odal fusion, named \textbf{M$^3$amba}. This unified framework benefits from the linear time complexity provided by Mamba \cite{gu2023mamba} and demonstrates improved transferability through CLIP image encoder and modality-specific adapters, as shown in the application transfer from the natural image domain to the remote sensing domain. Specifically, as shown in Fig. \ref{intro_fig}, we implant CLIP-driven modality-specific adapters into the encoder branch to introduce modality-specific semantic understanding with minimal training, thereby learning comprehensive visual semantic representations. This semantically guided cross-modal fusion enhances the understanding of global and local associative sequences, ensuring more robust and coherent feature representations. To achieve complete fusion, we introduce a three-branch state space model (SSM) that effectively interacts with CLIP semantic features through the proposed Cross-SS2D module to extract consistent and complementary information. This fusion process models the feature space of sequential data without the need for quadratic computational complexity, thus capturing inter-modal dependencies at minimal computational cost. While common methods perform pairwise fusion by training the encoder and fusion network from scratch, incomprehensive semantic information leads to incomplete fusion features and often redundant representations. Extensive experiments show that M$^3$amba outperforms current mainstream CNN, Transformer, and Mamba architectures in both performance and training efficiency. This reflects the great potential of the CLIP and Mamba dual-driven unified framework in multi-modal fusion tasks. To summarize, our contributions are three-fold:
\begin{itemize}
	\item[$\bullet$] We propose M$^3$amba, a novel end-to-end CLIP-driven Mamba model for effective and efficient multi-modal fusion. To our best knowledge, M$^3$amba is the first model that synergistically optimizes the powerful visual capabilities of CLIP and the efficient computational performance of Mamba.
	\item[$\bullet$] We design a multi-modal Mamba fusion architecture embedded with the proposed Cross-SS2D module, aiming to capture the comprehensive representation with linear complexity by enhancing the interactivity between modality mappings.
	\item[$\bullet$] Extensive experiments in the field of multi-modal remote sensing show that M$^3$amba surpasses SOTA methods in terms of effectiveness and training efficiency, demonstrating the potential of the CLIP and Mamba dual-driven multi-modal fusion framework.
\end{itemize}

\section{Related Works}
\subsection{Large-scale Image Representation Learning}
With the emergence of weakly labeled data at the scale of entire websites, we have witnessed a surge in new models for general visual representation learning. Concurrently, the scale of image models built through regular supervised learning has also been rapidly increasing \cite{zhai2022scaling, riquelme2021scaling}. To further enhance visual representation capabilities, researchers have begun focusing on contrastive learning and self-supervised learning with large datasets and large models. The success of BERT \cite{devlin2018bert} has sparked an emerging direction of constructing large-scale visual models using masked visual modeling \cite{he2022masked, xie2022simmim}. 
Vision-language models (VLMs), such as CLIP \cite{radford2021learning} and ALIGN \cite{jia2021scaling}, are trained on billion-scale, often noisy image-text datasets. These models consist of modality-specific encoders (image and text) that generate embeddings for each modality, therefore CLIP has been widely used as the vision backbone for its semantic representative feature and promising scalability even in the age of Multi-modal Large Language Models (MLLMs). 
Additionally, several methods apply the CLIP image encoder to 2D image variations to achieve better quality or controllability \cite{ye2023ip, ramesh2022hierarchical}. Some methods use the CLIP image encoder to extract features for rendering images \cite{mohammad2022clip, sanghi2022clip, xu2023dream3d}, while others adopt CLIP to extract semantically rich video representations \cite{luo2022clip4clip, luo2023towards, luo2024zero}. Approaches like LiT \cite{zhai2022lit} and BLIP-2 \cite{li2023blip} reduce the training costs of CLIP-like models by deploying pre-trained unimodal models.

Inspired by these developments, we utilize a pre-trained CLIP image encoder to extract semantic information from multi-modal images. However, its perception of a specific modality often suffers from understanding deviations due to domain gaps. Therefore, we introduce modality-specific adapters to further enhance intra-modal semantic understanding with minimal training overhead.

It is worth noting that in our implementation, we choose CLIP instead of ALIGN mainly based on the following considerations: 1) CLIP has stronger migration ability and more stable performance; 2) CLIP provides a more complete pre-trained model and interface; 3) Research in the field of remote sensing shows that CLIP is more suitable for processing the semantic features of remote sensing images \cite{dong2024changeclip}.

\subsection{Multi-modal Fusion in Remote Sensing}

Faced with the increasingly complex and diverse needs of scene representation applications, deep learning has made significant contributions to overcoming the technical bottleneck of multi-modal feature fusion \cite{mei2023lightweight,mei2023rotation}, and researchers have been advancing the study of multi-modal classification tasks in remote sensing, especially through CNN and Transformers \cite{roy2023multimodal, lu2023coupled, ding2024uncertainty, zhang2024multimodal, wang2023mutually, dong2023joint, wang2025s3f2net,mei2021hyperspectral,mei2022learning}. Although these methods integrate complementary information between different modalities, they are still limited in terms of computational efficiency and fusion completeness.
Recently, Mamba has emerged as a promising candidate for the next generation of base model backbones as it exhibits better scaling than Transformers while maintaining linear time complexity and has been developed in the field of multi-modal remote sensing. SpectralMamba \cite{yao2024spectralmamba} achieved improved performance by dynamically simplifying but fully modeling hyperspectral data in both spatial-spectral space and hidden state space. S$^2$Mamba \cite{wang2024s} mined spatial-spectral context features to achieve more efficient and accurate land cover analysis. MiM \cite{zhou2024mamba} introduced an innovative deployment architecture for hyperspectral image classification, improving the model's efficiency through a novel centralized Mamba Cross Scan (MCS) mechanism and T-Mamba encoder design.

Compared to these methods, M$^3$amba uses comprehensive semantic features adapted by CLIP to guide the fusion of multi-modal features with linear complexity, while Cross-SS2D ensures feature consistency and uses complementary information as gain.

\section{Methodology}

\begin{figure*}[t]
	\centering
	\includegraphics[width=1\linewidth]{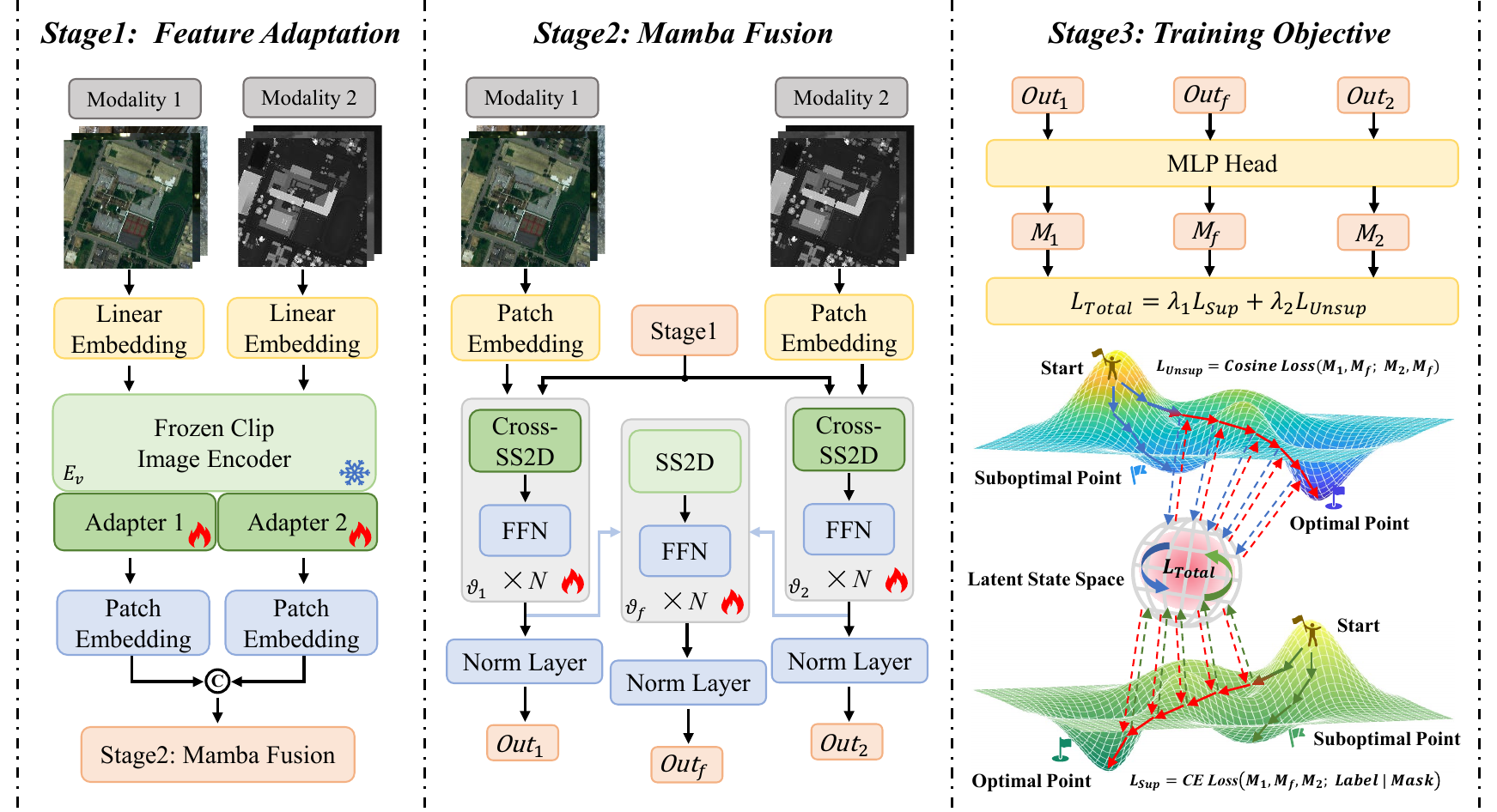}
	\vspace{-15pt}
	\caption{Overview of the M$^{3}$amba framework. For clarity, we split the end-to-end process of training into three stages: Feature Adaptation, Mamba Fusion with Cross-SS2D, and Training Objective. By utilizing the M$^{3}$amba framework to perform a feature-level fusion of the two modalities, we can apply the complete fusion features to different downstream tasks. MLP Head consists of several convolutional and linear layers, and CE stands for Cross Entropy.}
	\label{framework_fig}
	\vspace{-11pt}
\end{figure*}

\subsection{Problem Formulation}
In recent years, multi-modal data fusion has emerged as a crucial technique for enhancing the performance of various tasks, particularly in the field of remote sensing. In this paper, we focus on the application of multi-modal fusion to the pixel-level classification problem. This can be defined as accurately assigning each pixel in images from two modalities to the corresponding class. Given an image $X$ consisting of $q$ pixels, we aim to perform a pixel-level classification task using data from different modalities $X_1, X_2\in\mathbb{R}^{h\times w\times c}$. Both modalities learn features of the same scene with label information $L\in\mathbb{R}^{h\times w\times m}$, and the label for the $p$-th pixel can be represented as a one-hot vector $L^{q} = \{0^{m-1}, 1\}$, where $m$ denotes the number of classes. The objective of multi-modal pixel-level classification is to develop and train a model \( \psi(X_1, X_2) \) that effectively maps images from different modalities to the classification distribution \( C_{max}(X_1, X_2; L) \), indicating the probability that each pixel is associated with different classes under the guidance of the labels. A binary prediction map can be obtained through hard classification by a threshold \( \tau \) on the maximum probability for different classes.
\begin{equation}
	\psi\left(X_1,X_2\right)=\left\{\begin{matrix}0,&\text{if } C_{\max}\left(X_1,X_2; L\right)<\tau,\\1,&\text{otherwise.}\end{matrix}\right.
\end{equation}

Based on the above objectives, the logits can be expressed as \( C_{max}(X_1, X_2; L) \), specifically formulated as \( C_{max}(X_1, X_2; L) = \vartheta(X_1, X_2 \mid \beta_1, \beta_2) \). Here, we map the image feature space to the classification space using a nonlinear target model \( \vartheta(\cdot) \), where \( \beta_1 \) and \( \beta_2 \) represent the parameters of the two modality branches.

\subsection{Preliminaries}
\noindent\textbf{State Space Models}
State Space Models (SSMs) represent a class of sequence-to-sequence modeling systems characterized by constant dynamics over time, often used to represent linear time-invariant systems. 
Due to their linear complexity, SSMs can efficiently capture the inherent dynamics of a system through implicit mapping to latent states. 
Mathematically, an SSM is typically represented as a linear ordinary differential equation (ODE):
\begin{equation}
\frac{dh(t)}{dt} = Ah(t) + Bx(t), y(t) = Ch(t) + Dx(t),
\end{equation}
where \( x(t) \in \mathbb{R} \), \( h(t) \in \mathbb{R}^n \), and \( y(t) \in \mathbb{R} \) represent the input, hidden state, and output, respectively, with \( n \) being the state size. The remaining parameters include the state transition matrix \( A \in \mathbb{R}^{n \times n} \), projection parameters \( B \in \mathbb{R}^{n \times 1} \), \( C \in \mathbb{R}^{1 \times n} \), and skip connection \( D \in \mathbb{R} \).

To handle discrete sequences such as images and text, the ODE needs to be converted into a discrete function by introducing a predefined time scale parameter \( \Delta \in \mathbb{R}^D\). The discretization process of the above equations is as follows:
\begin{equation}
\overline{A} =\exp(\Delta A), \\
\overline{B} =(\Delta A)^{-1}(\exp(\Delta A)-I)\cdot\Delta B,
\end{equation}
\begin{equation}
\label{eq:4}
h^t=\overline{A}h^{t-1}+\overline{B}x^t, \\
y^{t}=Ch^{t} + Dx^{t},
\end{equation}
all matrices maintain the same dimensions across iterations of the operation. Furthermore, following Mamba, the matrix $\overline{B}$ can be approximated by a first-order Taylor series:
\begin{equation}
\overline{B}=(\exp(A)-I)A^{-1}B\approx(\Delta A)(\Delta A)^{-1}\Delta B=\Delta B.
\end{equation}


\noindent\textbf{2D-Selective-Scan Mechanism}
Due to the one-dimensional and causal properties of the Mamba selection mechanism, directly applying Mamba to visual tasks is unsuitable. 
For instance, while two-dimensional spatial information plays a crucial role in vision-related tasks, it is secondary in one-dimensional sequence modeling. Additionally, causal processing of input data prevents Mamba from absorbing information from parts of the data not yet scanned. 
These differences result in a limited receptive field, unable to capture potential correlations with unexplored patches. 
VMamba \cite{liu2024vmamba} introduces a Two-Dimensional Selective Scanning (SS2D) mechanism to solve this problem. 
SS2D transforms the input image into patch sequences along the horizontal and vertical axes, scanning in four directions: top-left to bottom-right, bottom-right to top-left, top-right to bottom-left, and bottom-left to top-right, generating independent sequences. This four-way scanning ensures that each element in the feature map contains information from all other positions in various directions. Consequently, it establishes a comprehensive global receptive field without linearly increasing computational complexity. The generated four sequences are then individually processed using selective SSM. Finally, all feature sequences are transformed back to their original 2D layout and merged to reconstruct the 2D feature map. SS2D is a core component of the Visual State Space (VSS) module, as shown in Figure \ref{framework_fig}, on which we build the hidden state space for cross-modal feature fusion.
\begin{algorithm}[tb]
\caption{Algorithm of Cross-SS2D.}
\label{alg:algorithm}
\textbf{Input}: Different modalities of data $X_{1}$, $X_{2}$.  \\
\textbf{Parameter}: Weight matrices $weight_{1}$, $bias_{1}$, $weight_{2}$, $bias_{2}$, $A_{1}$, $B_{1}$, $A_{2}$, $B_{2}$.  \\
\textbf{Output}: Fusion feature $H_{f}$. \\ 
\begin{algorithmic} [1]
 \STATE \textbf{Initial} $A_{1}$, $B_{1}$, $A_{2}$, $B_{2}$. \\
 \FOR{each block N = 1, 2, ..., 9}
 \STATE \textbf{1. Compute System Matrices}
 \STATE \quad $As_{1}$ $\leftarrow$ $-exp(A_{1})$ \quad $As_{2}$ $\leftarrow$ $-exp(A_{2})$
 \STATE \quad $Bs_{1}$ $\leftarrow$ $Split(einsum(X_{1}, weight_{1}) + bias_{1})$
 \STATE \quad $Bs_{2}$ $\leftarrow$ $Split(einsum(X_{2}, weight_{2}) + bias_{2})$
 \STATE \textbf{2. Cross Assignment}
 \STATE \quad $As_{f}$ $\leftarrow$ $(As_{1} + As_{2}) / 2$
 \STATE \quad $Bs_{1}$ $\leftrightarrow$ $Bs_{2}$
 \STATE \textbf{3. Extract Fusion Features}
 \STATE \quad $H_{f}$ $\leftarrow$ $selective\_scan(As_{f}, Bs_{1}, Bs_{2})$
 \ENDFOR
 \STATE \textbf{return} $H_{f}$ to the model.
 \end{algorithmic}
\end{algorithm}
\subsection{Network Architecture}
\noindent\textbf{Overview}
The overall structure of M$^{3}$amba, illustrated in Figure \ref{framework_fig}, is a multi-modal Mamba model built on top of a fixed CLIP image encoder, with images from different modalities used as inputs. 
Features corresponding to the same class from different modalities contain both consistent information and complementary insights. For instance, LiDAR provides enhanced spatial resolution, while hyperspectral data offers superior spectral resolution; infrared sensors excel at capturing thermal radiation, and visible light sensors are proficient at detailed texture data. However, a single sensor cannot capture all the complex details of an image, but the semantic information of images under the same scene is always consistent. Therefore, our goal is to \textbf{maximize the description of consistent information while leveraging complementary insights as benefits under comprehensive semantic feature guidance}.

Inspired by this, our proposed method aims to balance the informational discrepancies between the two modalities while utilizing their consistency and complementarity. 
For clarity, our network structure is conceptually divided into two parts, but it operates as a single end-to-end training process. Initially, modality-specific adapters are embedded in the pre-trained CLIP image encoder to capture comprehensive semantic information from different modalities. The output features, along with the input images, are then processed by Mamba's three-branch feed-forward network: \( \vartheta_1 \), \( \vartheta_2 \), and \( \vartheta_f \). 
Guided by an unsupervised consistency loss, the supervised fusion network with linear complexity is optimized to fully capture the features across modalities. 
Specifically, the fusion network consists of three main branches: two consistency branches \( \vartheta_1 \) and \( \vartheta_2 \), \( \vartheta_1 \) processes the feature sequence of the first modality (HSI), and \( \vartheta_2 \) processes the feature sequence of the second modality (LiDAR). They share the same network architecture, but use independent parameters to maintain the uniqueness of each modality feature, and map the unimodal and comprehensive semantic features to the hidden state space for interaction, so that complementary features and their consistency information can be thoroughly extracted. Additionally, the fusion branch \( \vartheta_f \) employs a selective scanning mechanism for global modeling, further integrating the complete fusion features. The entire fusion network undergoes \( N \) iterative updates, and the classification outputs of the three branches are finally used for loss function computation.

\noindent\textbf{Feature Adaptation}
By learning from a large number of image-text pairs, CLIP
can match images with their corresponding natural language descriptions. During training, the encoders are optimized by extracting features from input samples and aligning them in the embedding space using contrastive loss. 
This enables zero-shot classification and highlights CLIP's inherent advantage in extracting deep semantic visual feature representations. 
Leveraging this, we use the CLIP image encoder \(E_v\) to infer multi-modal input images \(X_1\) and \(X_2\), mapping them to the same feature space to reduce the domain gap between modalities and learn visual representations with rich semantics. In addition, in order to endow the original CLIP encoder with modality-specific perception capabilities for guiding more comprehensive feature fusion, we introduce modality-specific adapters. The resulting multi-modal features \(Y_1\) and \(Y_2\) have comprehensive modal scene semantic understanding and ensure more robust and coherent feature representations. This can be expressed as the equation:
\begin{equation}
	Y_1, Y_2 = E_v^{Adapter_1}(X_1), E_v^{Adapter_2}(X_2),
\end{equation}
where \( E_v^{Adapter_1} \) and \( E_v^{Adapter_2} \) represent the CLIP image encoder with modality-specific adapters added. The learnable adapters consist of two linear layers with residual connections, inserted after each attention block. We discard the last layer of the encoder because it is dedicated to classification. The resulting output features \( Y_1\) and \( Y_2\) are concatenated into \( Z \) after feature size transformation to provide guidance for the fusion network.

\noindent\textbf{Mamba Fusion}
As analyzed in the related work, previous multi-modal fusion methods excessively focused on the differing relationships between modalities, often neglecting the consistency across the data. Our fusion objective is to obtain complete fusion features between modalities, involving capturing local differences and overall invariants. 
By leveraging the obtained CLIP semantic feature \( Z \), we facilitate the fusion process to learn complete feature representations. Therefore, the entire encoding process can be expressed as:
\begin{gather}
	Out_1  =\vartheta_1(X_1,Z|\beta_1),\\
	Out_2  =\vartheta_2(X_2,Z|\beta_2), \\
	Out_f  =\vartheta_f(Out_1+Out_2|\beta_f),
\end{gather}
where \(Out_1\), \(Out_2\), and \(Out_f\) are the outputs of the three branches, and \(\beta_1\), \(\beta_2\), and \(\beta_f\) are their corresponding parameters. The fusion of features from both modalities significantly impacts the performance of downstream tasks. To that effect, designing a cross-modal complete attention mechanism becomes crucial. 

\noindent\textbf{Cross-SS2D}
As shown in Algorithm \ref{alg:algorithm}, the Cross-SS2D takes two features as input and generates a fused output while preserving the original shape of the features. The two inputs come from the comprehensive semantic features of CLIP and the unimodal image features, fully extracting independent features from each modality to complement the fused features and leveraging complementary insights to encompass comprehensive modal information. 
To achieve the model's context-aware capability, linear projection layers are employed to produce the system matrices \(B\), \(C\), and \(\Delta\) from the inputs. As specified in Equation \ref{eq:4}, matrices \(\overline{A}\) and \(\overline{B}\) encode the previous hidden state \(h^{t-1}\) and the input \(x^t\) to compute the current state \(h^t\).

Inspired by the cross-attention mechanisms \cite{chen2021crossvit} widely used in multi-modal tasks, we aim to facilitate the interaction of information between multiple modalities within the 2D selective scanning module. To achieve this goal, we use the \(\overline{B}\) matrix generated by the complementary modality in the selective scanning operation, enabling the SSM to provide complementary information for the current modality guided by the features from another modality. Additionally, the \(\overline{A}\) matrices corresponding to the two modalities are averaged to provide consistent information for the multi-modal input, ultimately resulting in the fusion feature \(H_f\). Specifically, this process can be expressed as:
\begin{gather}
	\overline{A}_{1}=\exp(\Delta_{1}A_{1}),\overline{A}_{2}=\exp(\Delta_{2}A_{2}),\\
	\overline{B}_{1}=\Delta_{1}B_{1},\overline{B}_{2}=\Delta_{2}B_{2},\\
	h_{1}^{t}=(\overline{A}_{1} + \overline{A}_{2})h_{1}^{t-1}/2+\overline{B}_{2}x_{1}^{t},\\
	H_f=y_{1}^t=C_{1}h_{1}^t+D_{1}x_{1}^t,
\end{gather}
where $\Delta$ is a predefined time scale parameter, \( x^t \) represents the input at time \( t \), and \( y^t \) denotes the selective scan output. Subscripts 1 and 2 represent unimodal features and CLIP features, respectively. To prevent redundant fusion, we choose $y_1^t$ as the fusion feature. \( \overline{A}_{1} \) and \( \overline{A}_{2} \) are used as the averaged state transition matrices providing multi-modal consistency, \( \overline{B}_{1} \) and \( \overline{B}_{2} \) are used as the cross-modal projection matrices providing complementary information. By implementing this Cross-SS2D fusion mechanism, we achieve a balanced and comprehensive representation of multi-modal features, leveraging both the complementary and consistent aspects of the input data. In addition, the A matrix enhances the feature consistency between modalities, making the fusion process smoother. The B matrix guides feature selection and avoids the transmission of redundant information. Through the interaction of the A matrix and the B matrix, Cross-SS2D can establish an efficient information transmission mechanism between modalities, improving the efficiency of feature selection and fusion between different modalities.
\begin{table*}[htbp]
	\centering
	\renewcommand{\arraystretch}{1.3}
	\setlength{\tabcolsep}{4mm}
	\caption{Ablation Studies on the Houston2013, the Augsburg, and the MUUFL Dataset. The Best Result is \textbf{Highlighted}}
	\begin{tabular}{c|ccc|ccc|ccc}
		\hline\hline
		& \multicolumn{3}{c|}{\textbf{Houston2013}} & \multicolumn{3}{c|}{\textbf{Augsburg}} & \multicolumn{3}{c}{\textbf{MUUFL}} \\
		\hline\hline
		\textbf{Scheme} & \textbf{OA}(\%) & \textbf{AA}(\%) & \textbf{$\kappa$}(×100) & \textbf{OA}(\%) & \textbf{AA}(\%) & $\kappa$(×100) & \textbf{OA}(\%) & \textbf{AA}(\%) & $\kappa$(×100) \\
		\hline
		(A) & 95.02 & 95.06 & 94.66 & 95.65 & 84.52 & 93.72 & 96.25 & 91.03 & 95.00 \\
		(B) & 95.84 & 95.84 & 95.55 & 96.27 & 86.68 & 94.66 & 97.16 & 94.70 & 96.22 \\
		(C) & 96.85 & 96.86 & 96.62 & 97.47 & 88.88 & 96.36 & 97.40 & 95.02 & 96.54 \\
		\textbf{(D)} & \textbf{97.31} & \textbf{97.32} & \textbf{97.11} & \textbf{98.19} & \textbf{93.32} & \textbf{97.39} & \textbf{97.84} & \textbf{96.10} & \textbf{97.13} \\
		\hline\hline
	\end{tabular} 
	\label{ablation_tab}
\end{table*}
\begin{table*}[htbp]
	\centering
	\renewcommand{\arraystretch}{1.3}
	\setlength{\tabcolsep}{3mm}
	\caption{Comparison Results on the Houston2013, the Augsburg, and the MUUFL Dataset. The Best Result is \textbf{Highlighted}}
	\begin{tabular}{c|c|ccc|ccc|ccc}
		\hline\hline
		\multirow{2}{*}{\textbf{Method}} & \multirow{2}{*}{\textbf{Model}} & \multicolumn{3}{c|}{\textbf{Houston2013}} & \multicolumn{3}{c|}{\textbf{Augsburg}} & \multicolumn{3}{c}{\textbf{MUUFL}} \\
		& &\textbf{OA}(\%) & \textbf{AA}(\%) & \textbf{$\kappa$}(×100) & \textbf{OA}(\%) & \textbf{AA}(\%) & $\kappa$(×100) & \textbf{OA}(\%) & \textbf{AA}(\%) & $\kappa$(×100) \\
		\hline\hline
		CALC$_{23}$ &CNN  &88.97  &90.78  &88.06  &91.73  &62.62  &87.98  &93.94  &74.09  &92.00  \\
		HCT$_{23}$ &CNN\&ViT  &91.15  &92.28  &90.40  &90.90  &65.24  &86.83  &92.95  &80.50  &90.69  \\
		UACL$_{24}$ &CNN  &95.37  &95.99  &95.00  &89.19  &74.30  &84.80  &88.29  &89.11  &84.78  \\
		Semi-ViT$_{22}$ & ViT  &85.46  &86.71  &86.71  &86.64  &64.67  &84.05  &92.46  &79.63  &89.49  \\
		MFT$_{23}$ & ViT &89.80  &91.51  &88.89  &90.49  &60.36  &86.26  &94.34  &81.48  &92.51  \\
		ExViT$_{23}$ & ViT  &91.40  &92.60  &90.66  &87.91  &63.20  &82.82  &94.37  &83.37  &92.54  \\
		MiM$_{24}$ &Mamba  &92.89  &94.21  &92.28  &88.63  &67.57  &86.56  &92.65  &82.22  &91.59  \\
		S$^{2}$Mamba$_{24}$ &Mamba  &93.36  &94.09  &92.79  &89.34  &70.41  &89.45  &94.19  &83.72  &91.98  \\
		\textbf{M$^3$amba(Ours)} & Mamba & \textbf{97.31} & \textbf{97.32} & \textbf{97.11} & \textbf{98.19} & \textbf{93.32} & \textbf{97.39} & \textbf{97.84} & \textbf{96.10} & \textbf{97.13} \\
		\hline\hline
	\end{tabular} 
	
	\label{hsi_tab}
	\vspace{-10pt}
\end{table*}
\subsection{Training Objective}
Given two modality inputs \( X_1^i \) and \( X_2^i \), our training strategy involves enhancing the supervised fusion network with linear complexity under the direction of an unsupervised loss to fully capture consistent and complementary features between modalities, where \( X_1^i \) represents the \( i \)-th sample of the first modality. For the supervised loss, we create an object mask map based on the ground truth, aiming to promote the consistency of the output probability distribution with the label \( L \) under the strong supervision constraint of the mask. The supervised loss calculation is as follows:
\begin{figure}[htbp]
	\centering
	\includegraphics[width=1\linewidth]{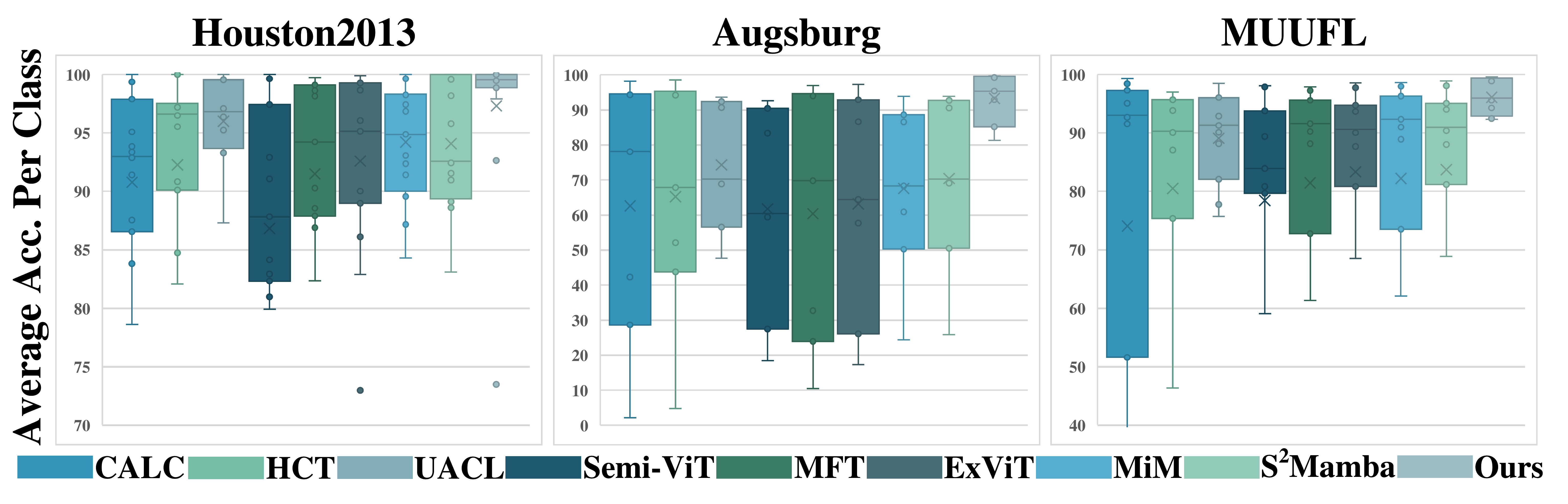}
	\caption{Comparison of box plots with other methods on three datasets.}
	\label{boxplot_fig}
\end{figure}
\begin{multline}
	\mathcal{L}_{Sup}=\sum_{i=1}^{X^{i}}\mathcal{L}_{CE}[(L,f_{1}\left(Out_1\right));
	(L,f_{2}\left(Out_2\right));\\
	(L,f_{f}\left(Out_f\right)) \ | \ Mask],
\end{multline}


where \( f_1(\cdot) \), \( f_2(\cdot) \), and \( f_f(\cdot) \) represent the MLP Head.
To minimize the differences between the unique and complete feature predictions, 
we calculate the sum of cross-modal cosine similarity losses of three different features for unsupervised learning. The proposed unsupervised consistency loss is calculated as follows:
\begin{multline}
	\mathcal{L}_{Unsup}=\sum_{i=1}^{X^{i}}[2-\cos(f_{1}\left(Out_1\right),f_{f}\left(Out_f\right))-\\
	\cos(f_{2}\left(Out_2\right),f_{f}\left(Out_f\right))].
\end{multline}

The complete loss function consists of supervised loss and unsupervised consistency loss. However, there is a consensus that strong supervised learning can provide more accurate guidance for model optimization. Therefore, we utilize supervised loss as the primary update and the unsupervised consistency loss as auxiliary optimization, as shown in the following equation: $\mathcal{L}_{Total} = \lambda_1\mathcal{L}_{Sup} + \lambda_2\mathcal{L}_{Unsup}$. In particular, $\lambda_1$ is usually greater than $\lambda_2$.


\begin{figure}[htbp]
	\centering
	\includegraphics[width=0.9\linewidth]{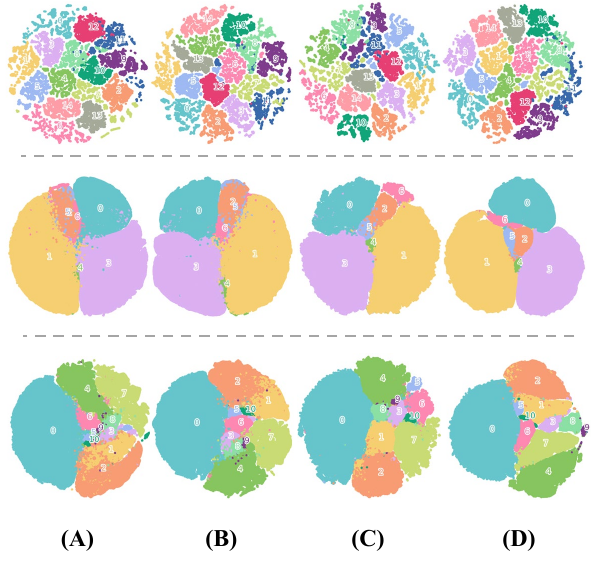}
	\caption{t-SNE for ablation on three datasets. The results from top to bottom correspond to Houston2013, Augsburg
		, and MUUFL datasets respectively.}
	\label{ablation_fig}
\end{figure}

\section{Experiments and Analysis}
\subsection{Experimental Settings}
\subsubsection{Datasets} 
To validate the effectiveness of the proposed method in analyzing multi-modal remote sensing data, we utilize three hyperspectral multi-modal datasets: the Houston2013 dataset \cite{debes2014hyperspectral}, the Augsburg dataset \cite{hong2021multimodal}, and the MUUFL dataset \cite{gader2013muufl} for remote sensing classification tasks.

\begin{table}[htbp]
	\small
	\renewcommand{\arraystretch}{1.3}
	\setlength{\tabcolsep}{1.5mm}
	\caption{A List of the Number of Training and Testing Samples for Each Class in Houston2013 Dataset}
	\label{tab-houstondataset}
	\centering
	\begin{tabular}{ccc|ccc}
		\hline\hline
		\textbf{Class No.}  & \textbf{Training} & \textbf{Testing} 
		&\textbf{Class No.}  & \textbf{Training} & \textbf{Testing} \\ 
		\hline\hline
		1  & 198 & 1053        &  9 & 193 & 1059 \\ 
		2  & 190 & 1064       &  10  & 191 & 1036 \\ 
		3  & 192 & 505         &	11  & 181 & 1054 \\ 
		4 & 188 & 1056                    & 12  & 192 & 1041 \\ 
		5  & 186 & 1056                    & 13  & 184 & 285 \\ 
		6  & 182 & 143                    &  14  & 181 & 247 \\
		7  & 196 & 1072            & 15  & 187 & 473 \\ 
		8  & 191 & 1053            &&& \\
		\hline\hline
	\end{tabular}
\end{table}
\begin{table}[htbp]
	\small
	\renewcommand{\arraystretch}{1.3}
	\setlength{\tabcolsep}{1.5mm}
	\caption{A List of the Number of Training and Testing Samples for Each Class in Augsburg Dataset}
	\label{Augsburg}
	\centering
	\begin{tabular}{ccc|ccc}
		\hline\hline
		\textbf{Class No.}  & \textbf{Training} & \textbf{Testing} 
		&\textbf{Class No.}  & \textbf{Training} & \textbf{Testing} \\ 
		\hline \hline
		1  & 146 & 13361         & 5  & 248 & 26609 \\
		2  & 7 & 1638           &  6 & 52 & 523 \\  
		3  & 264 & 30065         & 7  & 23 & 1507 \\  
		4  & 21 & 3830            &&&\\ 
		\hline\hline
	\end{tabular}
\end{table}
\begin{table}[t]
	\small
	\renewcommand{\arraystretch}{1.3}
	\setlength{\tabcolsep}{1.5mm}
	\caption{A List of the Number of Training and Testing Samples for Each Class in MUUFL Dataset}
	\label{MUUFL}
	\centering
	\begin{tabular}{ccc|ccc}
		\hline\hline
		\textbf{Class No.}  & \textbf{Training} & \textbf{Testing} 
		&\textbf{Class No.}  & \textbf{Training} & \textbf{Testing} \\ 
		\hline \hline
		1  & 1162 & 22084     & 7  & 112 & 2121 \\ 
		2  & 214 & 4056       & 8  & 312 & 5928 \\ 
		3  & 344 & 6538       &	9  & 69 & 1316 \\ 
		4  & 91 & 1735        & 10  & 9 & 174 \\
		5  & 334 & 6353       & 11  & 13 & 256 \\ 
		6  & 23 & 443          &&&  \\   
		\hline\hline
	\end{tabular}
\end{table}

\noindent\textbf{The Houston2013 Dataset}
The Houston2013 HSI-LiDAR Dataset is a widely used resource for remote sensing research, specifically designed for land cover classification tasks. It features hyperspectral imagery (HSI) captured by the ITRES CASI-1500 sensor over the University of Houston and nearby rural areas in Texas, USA. This dataset was initially collected in June 2012 and later made available for the IEEE GRSS Data Fusion Contest in 2013. The HSI comprises 144 spectral bands, spanning from 380 nm to 1050 nm, with a spatial resolution of 10 nm, while the LiDAR data provide a single-channel elevation map. Both data modalities share a spatial resolution of 2.5 meters, and the entire dataset consists of $349 \times 1905$ pixels. Predefined training and testing sets are available for classification tasks, with details in Table \ref{tab-houstondataset}.

\noindent\textbf{The Augsburg Dataset}
The Augsburg HS-SAR-LiDAR Dataset originates from a multi-modal data collection effort in Augsburg, Germany, and includes hyperspectral, synthetic aperture radar (SAR), and digital surface model (DSM) images. The hyperspectral imagery was collected using the HySpex sensor, delivering data across 180 spectral bands between 400 nm and 2500 nm. The selected study area includes 332,485 pixels with a ground sampling distance (GSD) of 30 meters. Additionally, the dataset provides four SAR bands and DSM elevation data, with the ground truth derived from manual labeling. This dataset is ideal for urban area classification, covering various land cover types. The details are outlined in the Table \ref{Augsburg}.

\noindent\textbf{The MUUFL Dataset}
The MUUFL Gulfport Scene Dataset was gathered in November 2010 at the Gulf Park Campus of the University of Southern Mississippi in Long Beach, Mississippi. The dataset consists of hyperspectral images, captured by the CASI-1500 sensor, with 64 available spectral bands ranging from 375 nm to 1050 nm. The spatial resolution is approximately 0.54 by 1.0 meters. The dataset also includes LiDAR elevation data from two grates, providing further information for terrain analysis. The scene covers 11 distinct land cover categories, with labeled pixels available for training and testing. The details are outlined in the Table \ref{MUUFL}.


\begin{table}[htbp]
	\centering
	\small
	\setlength{\tabcolsep}{0.8mm}
	\renewcommand{\arraystretch}{1.3}
	\caption{Average Training Time Analysis on Three Datasets. The Best Result is \textbf{Highlighted}}
	\begin{tabular}{c|ccccccc}
		\hline\hline
		& Ours & MiM &ExViT & MFT & Semi-ViT & HCT & CALC \\
		\hline\hline
		Train. (min) &\textbf{17.67}  &18.71 &44.03  &42.62  &50.92 &24.28 &29.15    \\
		AA (\%) &\textbf{96.10}  &82.22 &83.37 &81.48 &79.63  &80.50  &74.09  \\
		\hline\hline
	\end{tabular} 
	
	\label{computation_cost_cls_tab}
\end{table}
\begin{table}[!h]
	\centering
	\small
	\setlength{\tabcolsep}{3mm}
	\renewcommand{\arraystretch}{1.3}
	\caption{Ablation Study of Feature Adaptation with Average Results on Three Datasets}
	\begin{tabular}{c|cccc}
		\hline\hline
		& OA(\%) & AA(\%) & $\kappa$($\times 100$) & Train. (min) \\
		\hline\hline
		Infer      &  95.23  &  90.19  &  94.17  & 15.82     \\
		Fine-tune  &  98.10  &  95.36  &  97.40  & 46.08     \\
		Ours       &  97.78  &  95.58  &  97.21  & 17.67    \\
		\hline\hline
	\end{tabular} 
	
	\label{ablation_feature_adaptation}
\end{table}

\subsubsection{Evaluation metrics}
To evaluate the performance in classification, we employ three metrics: Overall Accuracy (OA), Average Accuracy (AA), and the kappa ($\kappa$). OA measures the ratio of correctly classified samples to the total number of samples. AA represents the average accuracy across all classes. The $\kappa$ coefficient is a statistical
metric assessing the agreement between the classification map generated by the model and the ground truth.

\subsubsection{Implement details}
All experiments are conducted on a server equipped with an NVIDIA A100 Tensor Core GPU. For hyperspectral remote sensing images, due to their dense objects and high resolution, the data samples are usually cropped to a size of 32 × 32 and use patch size 1. For all datasets, we use ViT-B/16 as the backbone of CLIP,
and the training process employs the AdamW optimizer with an initial learning rate set to 1e-4, a weight decay of 1e-3, and a batch size of 8, spanning 200 epochs.

\begin{table*}[htbp]
	\small
	\renewcommand{\arraystretch}{1.3}
	\centering
	\setlength{\tabcolsep}{1.1mm}{
		\caption{OA, AA and Kappa Coefficient on the Houston2013 Dataset by Considering HSI and LiDAR Data. The Best Result is \textbf{Highlighted}. H and L Respectively Indicate that Our Method is Trained Using Only HSI or LiDAR Data.}
		\label{tab-vsh}    
		\begin{tabular}{c|c|c|c|c|c|c|c|c|c|c|c|c}
			\hline
			\hline
			~ & \textbf{Class Name} & \textbf{CALC$_{23}$} & \textbf{HCT$_{23}$} & \textbf{UACL$_{24}$} & \textbf{Semi-ViT$_{22}$} & \textbf{MFT$_{23}$} & \textbf{ExViT$_{23}$} & \textbf{MiM$_{24}$}&\textbf{S$^{2}$Mamba$_{24}$} & \textbf{M$^3$amba} & \textbf{H} & \textbf{L}\\ 
			\hline
			\hline
			1 & Healthy grass & 78.63  & 97.34  & 93.68  & 82.59  & 82.34  & 82.91  & 94.89  & 83.10  & \textbf{100.00} & 95.19  & 89.73   \\ 
			2 & Stressed grass  & 83.83  & 96.62  & 97.10  & 82.33  & 88.78  & 98.68  & 98.27  & \textbf{100.00}  & \textbf{100.00} & 98.13  & 84.66   \\
			3 & Synthetic grass  & 93.86  & 84.75  & 99.84  & 97.43  & 98.15  & 99.60  & \textbf{100.00}  & 99.60  & \textbf{100.00}  & 92.12  & 91.67  \\
			4 & Tree  & 86.55  & 96.78  & 96.82  & 92.93  & 94.35  & 99.15  & 96.87  & 98.20  & \textbf{100.00} & 79.58  &  81.32 \\
			5 & Soil  & 99.72  &\textbf{100.00}  & 99.55  & 99.84  & 99.12  & 99.91  & 99.86  & \textbf{100.00}  & 99.90  & 97.13  & 88.29  \\
			6 & Water  & 97.90  & 96.50  & 96.82  & 84.15  & 99.30  & 99.30  & \textbf{99.65}  & 95.80  & 92.65 & 80.45  & 73.59   \\
			7 & Residential  & 91.42  & 82.09  & 93.29  & 87.84  & 88.56  & 96.08  & 91.42  & 89.37  & \textbf{99.81} & 81.94  & 83.57   \\
			8 & Commercial  & 92.88  & 95.54  & 87.31  & 79.93  & 86.89  & 90.03  & 87.17  & 88.60  & \textbf{97.93} & 88.39  &  87.48  \\
			9 & Road  & 87.54  & 90.84  & 93.66  & 82.94  & 87.91  & 86.12  & 84.31  & 92.45  & \textbf{98.87} & 81.47  & 75.84   \\
			10 & Highway  & 68.53  & 58.88  & 95.26  & 52.93  & 64.70  & 72.97  & 92.38  & 92.57  & \textbf{99.48}& 89.39  & 86.53    \\
			11 & Railway  & 93.36  & 97.53  & 96.39  & 80.99  & 98.64  & 88.99  & 90.03  & 91.56  & \textbf{99.20} & 87.47  & 83.07  \\
			12 & Park lot 1  & \textbf{95.10}  & 90.11  & 93.52  & 91.07  & 94.24  & 90.39  & 89.58  & 90.97  & 73.50 & 70.18  &  52.95  \\
			13 & Park lot 2  & 92.98  & 97.19  & 97.13  & 87.84  & 90.29  & 90.18  & 93.08  & 89.12  & \textbf{99.25} & 86.63  & 67.32   \\
			14 & Tennis court  & \textbf{100.00}  & \textbf{100.00}  & \textbf{100.00}  & \textbf{100.00}  & 99.73  & 99.60  & 97.44  & \textbf{100.00}  & 99.69 & 91.19  & 84.25   \\
			15 & Running track  & 99.37  & \textbf{100.00}  & 99.56  & 99.65  & 99.58  & 95.14  & 98.33  & \textbf{100.00}  & 99.57  & 96.25  & 84.61  \\
			\hline
			~ & OA(\%)  & 88.97  & 91.15  & 95.37  & 85.46  & 89.80  & 91.40  & 92.89  & 93.36  & \textbf{97.31}  & 86.59  &  79.97 \\
			~ & AA(\%)  & 90.78  & 92.28  & 95.99  & 86.71  & 91.51  & 92.60  & 94.21  & 94.09  & \textbf{97.32} & 87.70  &  80.99  \\
			~ & $\kappa$($\times 100$)  & 88.06  & 90.40  & 95.00  & 86.71  & 88.89  & 90.66  & 92.28  & 92.79  & \textbf{97.11} & 86.33  &  80.71  \\
			\hline
			\hline
	\end{tabular}}
\end{table*}
\begin{figure*}[htbp]
	\centering
	\includegraphics[width=1\linewidth]{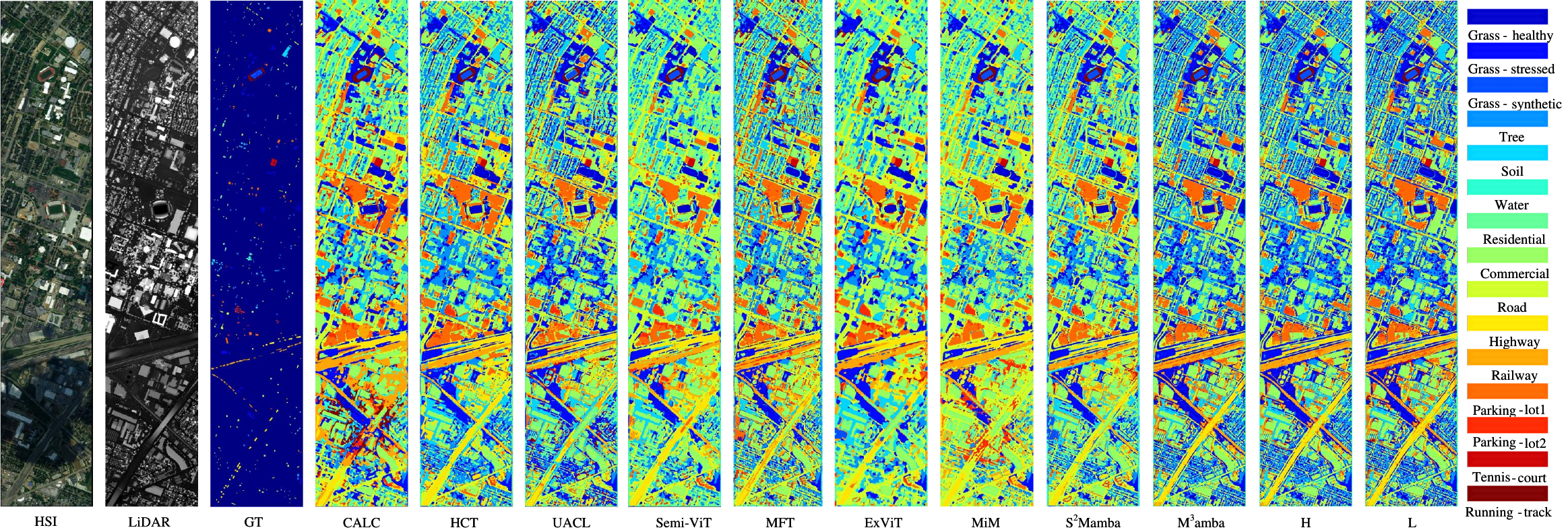}
	\centering
	\caption{Visualization of false-color HSI and LiDAR images using different comparison methods based on the Houston2013 dataset. H and L respectively indicate that our method is trained using only HSI or LiDAR data.}
	\vspace{-0.1in}
	\label{houston2013}
\end{figure*}

\subsection{Ablation Study}
\subsubsection{Ablation analysis of different components}
To evaluate the individual contributions of different components in our proposed method, we conduct ablation studies on three remote sensing datasets. Specifically, we design four schemes: (A) directly concatenating multi-modal images to replace the output of the feature adaptation stage; (B) not using our Cross-SS2D module and using the original SS2D module to process both inputs and concatenate the output to the fusion branch; (C) removing unsupervised consistency loss; (D) the complete M$^3$amba. The results of the ablation study are shown in Table \ref{ablation_tab}. We observe that M$^3$amba achieves optimal performance across the three datasets. In scheme (A), directly using the original image information to guide the fusion leads to an average OA reduction of 2.14\% on the three datasets, because the modality adapter reduces the inter-domain differences between different modalities, making each modality more accurate in specific tasks. When this module is removed, the fusion process of the modalities is affected, resulting in insufficient information interaction, which affects the overall classification effect. The effectiveness of our cross-attention fusion module Cross-SS2D is demonstrated in (B). Without the Cross-SS2D fusion module, OA, AA, and $\kappa$ decreased by 1.36\%, 3.17\%, and 1.73\% on average, respectively. This is because Cross-SS2D improves the synchronization and complementarity between modal features through the interaction of selective scanning and global information. After removing this module, the model cannot effectively fuse the information of the two modalities, resulting in a decrease in classification accuracy. The removal of the unsupervised consistency loss in scheme (C) also led to a significant decrease in performance. This is because the unsupervised consistency loss makes the features between the two modalities more matched by promoting the consistency of cross-modal features. After removing this loss function, the complementary information between the modalities cannot be effectively synchronized, resulting in inconsistency in information between the modalities, which affects the classification accuracy after fusion. In summary, our proposed CLIP feature adaptation and Mamba fusion method can perform joint optimization and effectively improve the performance of multi-modal learning. Additionally, to more intuitively demonstrate the impact of different components on performance, 
we use t-SNE to visualize the final classification results, as shown in Figure \ref{ablation_fig}.
The visualization results of different schemes have the same trend as the results in the table. In general, our M$^3$amba achieves complete feature fusion under the guidance of rich semantics. Comprehensive ablation experiments confirm that M$^3$amba achieves optimal results in both accuracy and visualization.

\begin{table*}[htbp]
	\small
	\renewcommand{\arraystretch}{1.3}
	\centering
	\setlength{\tabcolsep}{1.1mm}{
		\caption{OA, AA and Kappa Coefficient on the Augsburg Dataset by Considering HSI and LiDAR Data. The Best Result is \textbf{Highlighted}. H and L Respectively Indicate that Our Method is Trained Using Only HSI or LiDAR Data.}
		\label{tag-Augsburg}	
		\begin{tabular}{c|c|c|c|c|c|c|c|c|c|c|c|c}
			\hline\hline
			~ & \textbf{Class Name} & \textbf{CALC$_{23}$} & \textbf{HCT$_{23}$} & \textbf{UACL$_{24}$} & \textbf{Semi-ViT$_{22}$} & \textbf{MFT$_{23}$} & \textbf{ExViT$_{23}$} & \textbf{MiM$_{24}$}&\textbf{S$^{2}$Mamba$_{24}$} & \textbf{M$^3$amba} & \textbf{H} & \textbf{L} \\ 	
			\hline\hline
			1 & Forest & 94.34  & 94.23  & 93.66  & 90.41  & 94.65  & 92.89  & 93.91  & 93.90  & \textbf{99.63} & 96.13 &  96.88  \\
			2 & Commercial Area  & 98.24  & 98.54  & 90.70  & 92.64  & 96.90  & 97.28  & 88.61  & 90.55  &\textbf{99.24} & 91.59 & 64.10  \\
			3 & Residential Area  & 78.07  & 43.79  & 68.91  & 60.41  & 69.80  & 64.44  & 60.93  & 69.09  & \textbf{85.19} & 78.32 & 82.71   \\
			4 & Industrial Area  & 94.57  & 95.33  & 92.38  & 83.40  & 93.98  & 86.63  & 86.55  & 92.70  & \textbf{99.60} & 97.63 & 55.17   \\
			5 & Low Plants  & 28.68  & 67.88  & 56.58  & 59.41  & 32.70  & 57.74  & 68.29  & 70.30  & \textbf{95.39}  & 79.59 & 61.35  \\
			6 & Allotment  & 2.20  & 4.82  & 47.63  & 18.44  & 10.52  & 17.28  & 24.46  & 25.84  & \textbf{81.29} & 70.23 & 29.65   \\
			
			7 & Water  & 42.27  & 52.09  & 70.26  & 27.51  & 23.98  & 26.14  & 50.24  & 50.49  & \textbf{92.89} & 85.48 & 63.76   \\
			\hline
			~ & OA(\%)  & 91.73  & 90.90  & 89.19  & 86.64 & 90.49  & 87.91  & 88.63  & 89.34 & \textbf{98.19} & 89.36 & 77.11   \\
			~ & AA(\%)  & 62.62  & 65.24 & 74.30  & 64.67 & 60.36  & 63.20  & 67.57  & 70.41 & \textbf{93.32} &85.57  &  64.80  \\
			~ & $\kappa$($\times 100$)  & 87.98  & 86.83  & 84.80  & 84.05 & 86.26  & 82.82  & 86.56  & 89.45 & \textbf{97.39} & 88.10 &  78.79  \\
			\hline\hline
	\end{tabular}}
\end{table*}
\begin{figure*}[htbp]
	\centering
	\includegraphics[width=1\linewidth]{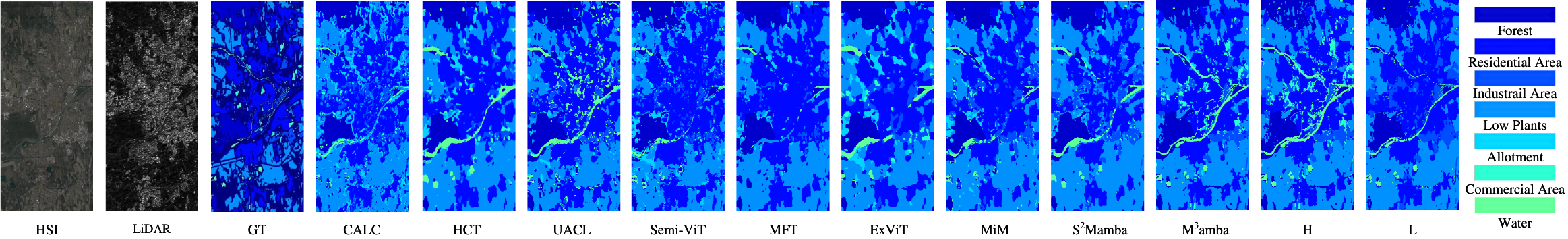}
	\centering
	\caption{Visualization of false-color HSI and LiDAR images using different comparison methods based on the Augsburg dataset. H and L respectively indicate that our method is trained using only HSI or LiDAR data.}
	\label{Augsburg_2.0}
\end{figure*}

\subsubsection{Ablation analysis of feature adaptation}
To further explore the effectiveness of semantically guided fusion through feature adaptation, we set up three schemes, as shown in Table \ref{ablation_feature_adaptation}. We use ViT-B/16 as the backbone of CLIP. Infer refers to directly using the pre-trained CLIP image encoder to infer multi-modal images, and Fine-tune refers to fine-tuning the CLIP image encoder. We introduce modality-specific adapters to train only a very small number of parameters. It can be observed that direct inference can bring a small improvement in training efficiency, but the lack of modality-specific scene understanding leads to a significant decrease in accuracy. This problem can be alleviated by fine-tuning, but the time cost is high. Our method aims to balance training efficiency and the performance of downstream tasks, achieving the best balance.

\begin{table*}[htbp]
	\small
	\renewcommand{\arraystretch}{1.3}
	\centering
	\setlength{\tabcolsep}{0.9mm}{
		\caption{OA, AA and Kappa Coefficient on the MUUFL Dataset by Considering HSI and LiDAR Data. The Best Result is \textbf{Highlighted}. H and L Respectively Indicate that Our Method is Trained Using Only HSI or LiDAR Data.}
		\label{tab-M}	
		\begin{tabular}{c|c|c|c|c|c|c|c|c|c|c|c|c}
			\hline\hline
			~ & \textbf{Class Name} & \textbf{CALC$_{23}$} & \textbf{HCT$_{23}$} & \textbf{UACL$_{24}$} & \textbf{Semi-ViT$_{22}$} & \textbf{MFT$_{23}$} & \textbf{ExViT$_{23}$} & \textbf{MiM$_{24}$}&\textbf{S$^{2}$Mamba$_{24}$} & \textbf{M$^3$amba} & \textbf{H} & \textbf{L}\\ 
			
			\hline\hline
			1 & Trees  & 97.31  & 97.03  & 91.29  & 97.89  & 97.90  & 98.58  & 98.64  & 98.93  & \textbf{99.57} & 91.13 & 96.79   \\
			2 & Grass-Pure  & \textbf{93.00}  & 90.29  & 82.06  & 79.71  & 92.11  & 87.70  & 92.85  & 88.05  & 92.50  & 87.04 & 86.07 \\
			3 & Grass-Groundsurface  & 91.57  & 90.07  & 77.79  & 84.67  & 91.80  & 90.96  & 92.54  & 91.31  & \textbf{96.18} & 90.60 & 89.35  \\
			4 & Dirt-And-Sand  & 95.10  & 94.18  & 90.13  & 89.40  & 91.59  & 90.61  & 92.33  & 90.96  & \textbf{95.95} & 93.79 & 81.41  \\
			5 & Road-Materials  & 95.91  & 93.86  & 88.16  & 93.81  & 95.60  & 94.73  & 96.34  & 95.08  & \textbf{98.88} & 94.78 & 85.97  \\
			6 & Water  &99.32  & 95.71  & 98.50  & 81.04  & 88.19  & 93.68  & 88.93  & 94.03  & \textbf{99.43}  & 92.75 & 80.19 \\
			7 & Buildings’Shadow  & \textbf{92.69} & 87.09  & 91.45  & 83.92  & 90.27  & 90.05  & 91.01  & 90.40  & 92.32  & 80.10 & 72.91 \\
			8 & Buildings & 98.45  & 96.61  & 92.91  & 98.11  & 97.26  & 97.76  & 98.00  & 98.11  & \textbf{99.44} & 97.40 & 99.32  \\
			9 & Sidewalk &51.60  & 46.35  & 75.71  & 59.12  & 61.35  & 68.54  & 62.09  & 68.89  & \textbf{92.92}& 85.37 & 51.32   \\
			10 & Yellow-Curb  & 0.00  & 18.97  & \textbf{96.07}  & 14.37  & 17.43  & 23.56  & 18.17  & 23.91  & 94.30  & 84.65 & 11.49 \\
			11 & ClothPanels  & 0.00  & 75.39  & \textbf{96.18}  & 80.86  & 72.79  & 80.86  & 73.53  & 81.21  & 95.61  & 90.19 & 78.93 \\
			\hline
			~ & OA(\%)  & 93.94  & 92.95  & 88.29  & 92.46  & 94.34  & 94.37  & 92.65  & 94.19  & \textbf{97.84}& 88.91 &  84.16  \\
			~ & AA(\%)  & 74.09  & 80.50  & 89.11  & 79.63  & 81.48  & 83.37  & 82.22  & 83.72  & \textbf{96.10} &89.80  & 75.80  \\
			~ & $\kappa$($\times 100$)  & 92.00  & 90.69  & 84.78  & 89.49  & 92.51  & 92.54  & 91.59  & 91.98  & \textbf{97.13} & 85.75 & 83.78  \\
			\hline\hline
	\end{tabular}}
\end{table*}
\begin{figure*}[htbp]
	\centering
	\includegraphics[width=1\linewidth]{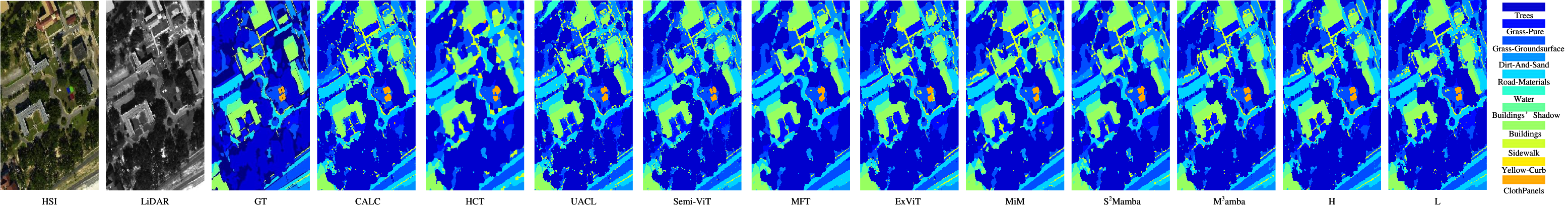}
	\centering
	\caption{Visualization of false-color HSI and LiDAR images using different comparison methods based on the MUUFL Gulfport scene dataset. H and L respectively indicate that our method is trained using only HSI or LiDAR data.}
	\label{muufl_2.0}
\end{figure*}

\subsection{Comparisons with Previous Methods}
To better demonstrate the superiority of our method, we conduct a comprehensive performance comparison with recent advanced multi-modal fusion methods, including CNN-based structures CALC \cite{lu2023coupled}, HCT \cite{zhao2022joint}, UACL \cite{ding2024uncertainty}; Transformer-based structures Semi-ViT \cite{cai2022semi}, MFT \cite{roy2023multimodal}, ExViT \cite{yao2023extended}; and Mamba-based structures MiM \cite{zhou2024mamba}, S$^2$Mamba \cite{wang2024s}. The subscript of each method in the comparative experiment table represents the publication time of this method, for example, the subscript 24 means it was published in 2024. The average results are summarized in Table \ref{hsi_tab}, with our model achieving significant improvements in overall accuracy (OA), average accuracy (AA), and the kappa ($\kappa$) coefficient across all datasets. 
In the Houston2013 dataset, as shown in Table \ref{tab-vsh}, M$^3$amba achieves the highest overall performance, which is a significant improvement over the closest competitor, UACL. The superior performance of M$^3$amba can be attributed to its CLIP-driven modality-specific adapters, which allow for better cross-modal semantic understanding compared to CNN-based methods. This is especially important in this dataset, where high spectral resolution and spatial detail are critical for accurate classification. For example, M$^3$amba achieves perfect classification accuracy for similar categories like Stressed grass and Synthetic grass, which often pose a challenge in terms of distinguishing subtle spectral differences. The CLIP-driven modality-specific adapters ensure that the model can better differentiate these two categories by refining the feature representations of both HSI and LiDAR modalities. This highlights how M$^3$amba’s ability to capture complementary information from different modalities—thanks to its Cross-SS2D module—improves classification accuracy for fine-grained classes. Additionally, M$^3$amba demonstrates robust performance across nearly all land cover classes, but its significant improvement is particularly noticeable in categories like Highway and Road, where the fusion of spectral and spatial information provided by the model enhances feature extraction and reduces misclassification.

The Augsburg dataset presents a more challenging scenario, but M$^3$amba again surpasses all competing methods with an OA of 98.19\%, an AA of 93.32\%, and a kappa coefficient of 97.39\%, as shown in Table \ref{tag-Augsburg}. One key reason for this outstanding performance is the ability of the Cross-SS2D module to maintain feature integrity across different data modalities. For example, in the Forest category, M$^3$amba outperforms competitors by leveraging both the spectral details from HSI and the structural information from LiDAR, which is crucial for distinguishing different types of vegetation. Similarly, in Industrial areas, where spatial complexity and high variability in land cover types are prevalent, M$^3$amba shows consistent improvements by modeling the interactions between spectral and spatial features more effectively than the CALC model.
The MUUFL dataset contains a mix of urban and natural land cover categories, as shown in Table \ref{tab-M}, M$^3$amba outperforms all other methods, achieving an OA of 97.84\%, an AA of 96.10\%, and a kappa coefficient of 97.13\%. Compared to the MiM model, which attains an OA of 92.65\%, M$^3$amba offers a 5.19\% improvement. In particular, M$^3$amba performs well in the sidewalk class, achieving an accuracy of 92.92\%, far exceeding other comparison methods. These results highlight the effectiveness of the Cross-SS2D module in maintaining the integrity of fine-grained features across different data modalities and its ability to model long sequences, which is a key advantage over other Mamba-based methods such as MiM and S$^2$Mamba. In the Buildings category, the model’s performance is also significantly enhanced. The fusion of HSI and LiDAR data provides a richer representation of building structure and surface features, which is often difficult to achieve with single-modal approaches. The model’s ability to effectively leverage both spatial and spectral features allows it to better capture complex building shapes and variations in urban environments.

To investigate the specific contribution of the complementarity of different modal data to the fusion performance, we train M$^3$amba using only HSI or LiDAR data and compare it with the results of multi-modal data training. The results in Tables \ref{tab-vsh}, \ref{tag-Augsburg}, \ref{tab-M} show that when combining the two data modalities, the spectral advantages of HSI and the spatial structure advantages of LiDAR can be simultaneously exploited to improve classification accuracy in complex scenes. For example, although the spectral features of objects such as buildings may be similar in HSI, the elevation information of LiDAR helps to accurately identify these objects. Objects with obvious spectral differences, such as water and soil, have significantly improved accuracy in the fused classification results. We can also find that the complementarity of the two modalities allows some categories to reach 100\% accuracy after fusion, which does not happen when using only single modal data.

Furthermore, as shown in Figure \ref{boxplot_fig}, the box plot of our method is taller and more compact than the competing methods, indicating that our method achieves robust detection performance for each class. This effectively mitigates the issue of low detection accuracy for individual classes observed in the competing methods.

In addition to the accuracy metrics, our method also demonstrates a notable reduction in training time compared to other models, as shown in Table \ref{computation_cost_cls_tab}. Our method achieves the lowest average training time across the three datasets, outperforming CNN-based, Transformer-based, and Mamba-based architectures and far exceeds all other methods in terms of average accuracy. This is attributed to the design of our efficient fusion architecture, in stark contrast to the quadratic complexity in traditional Transformer models. The combination of efficient feature fusion and comprehensive semantic guidance makes M$^3$amba both an accurate and computationally efficient model for multi-modal remote sensing tasks.



\subsection{Analysis of Failure Cases}
Although M$^3$amba achieves the highest accuracy on multiple categories in multiple datasets, there are still cases of poor classification, especially in complex scenes, where some categories perform poorly, partly due to the ambiguity of features or spectral overlap. Through the analysis of failed cases, we find the following two points. Texture fuzzy categories: such as Water (in Table \ref{tab-vsh}), because its texture features are similar to the surrounding environment in hyperspectral images, lead to classification errors. Especially in LiDAR data, these objects lack significant elevation differences, which affects the judgment of the model. Complex scene problems: In densely built-up areas in cities, objects such as parking lots (in Table \ref{tab-vsh}) lack clear spatial structure and have similar spectral features to the surrounding environment (such as roads and buildings), resulting in reduced classification accuracy.

\subsection{Result Visualization}
We visualize the results by assigning a unique color to each class. Figure \ref{houston2013}, Figure \ref{Augsburg_2.0}, and Figure \ref{muufl_2.0} demonstrate the excellent performance of our method on the full remote sensing image dataset, and our visualization is still better than most of the comparison methods even when trained with only single-modal data. M$^3$amba extracts fundamental properties of multi-modal images by establishing comprehensive fused features and invariant representations. It utilizes CLIP to generate comprehensive semantic features to guide the efficient fusion of multi-modal features. This enhances the model's generalization ability and preserves complete information between modalities, resulting in richer and more detailed representations in the classification maps. Overall, M$^3$amba is highly suitable for generating more robust and intricate detailed multi-modal classification maps.


\section{Conclusion}
In this paper, we introduce M$^3$amba, a novel end-to-end CLIP-driven Mamba unified framework for multi-modal fusion, combining CLIP's powerful multi-modal visual semantic representation capabilities with the efficiency of SSM. By introducing CLIP-driven modality-specific adapters and the Mamba fusion architecture embedded with the cross-attention module Cross-SS2D, M$^3$amba addresses the primary challenges of existing methods, including high computational complexity, limited generalization, and incomplete feature fusion, achieving optimal performance in both accuracy and efficiency. Extensive evaluations on multiple remote sensing multi-modal datasets demonstrate the generalizability and interpretability of M$^3$amba. This work showcases the tremendous potential for future research in efficient and effective multi-modal learning, highlighting the potential of combining pre-trained multi-modal models with state space modeling techniques to advance the field of multi-modal fusion. In addition, due to the efficient training and excellent performance of M$^3$amba, in future work, we will focus on verifying the effectiveness of M$^3$amba in specific application scenarios, especially in multiple practical scenarios such as agricultural monitoring, urban planning, and ecological protection, as well as exploring new directions such as dynamic target detection and real-time multi-modal data processing, to further demonstrate its wide applicability and strong potential in practical remote sensing tasks.

	\ifCLASSOPTIONcaptionsoff
\newpage
\fi
{
	\bibliographystyle{IEEEtran}
	\bibliography{refs.bib}
}


\begin{IEEEbiography}[{\includegraphics[width=1in,height=2in,clip,keepaspectratio]{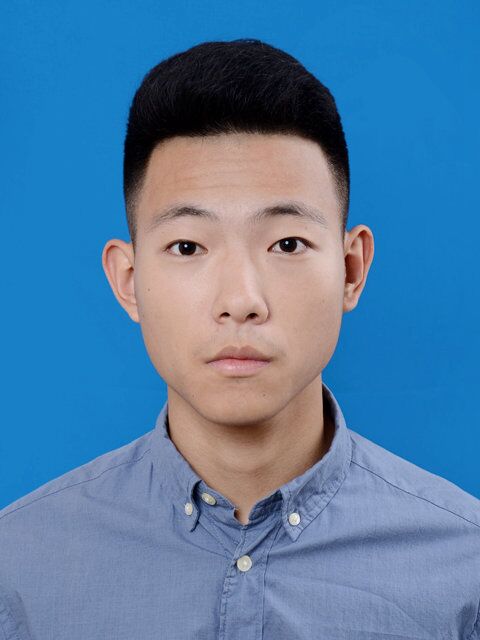}}]{Mingxiang Cao} received the B.E. degree in Telecommunications Engineering from Xidian University, Xi’an, China in 2023. He is currently pursuing the M.S. degree with the Image Coding and Processing Center at State Key Laboratory of Integrated Service Network, Xidian University, Xi'an, China. His research interests include multimodal image processing, remote sensing classification, and object detection.
\end{IEEEbiography}
\begin{IEEEbiography}[{\includegraphics[width=1in,height=2in,clip,keepaspectratio]{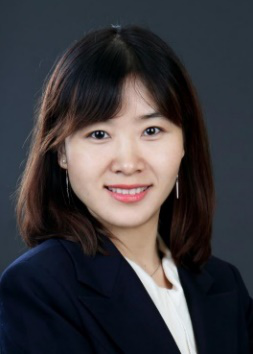}}]{Weiying Xie}(Senior Member, IEEE) received the Ph.D. degree in communication and information systems from Xidian University, Xi’an, China, in 2017. Currently, she is a Professor with the State Key Laboratory of Integrated Services Networks, Xidian University. She has published over 50 articles in refereed journals and proceedings, including IEEE Transactions on Image Processing, IEEE Transactions on Geoscience and Remote Sensing, IEEE Transaction on Neural Networks and Learning Systems, IEEE Transactions on Cybernetics, and Conference on IEEE/CVF Computer Vision and Pattern Recognition Conference (CVPR) and Association for the Advancement of Artificial Intelligence (AAAI). Her research interests include neural networks, machine learning, hyperspectral image processing, and high-performance computing.	
\end{IEEEbiography}
\begin{IEEEbiography}[{\includegraphics[width=1in,height=2in,clip,keepaspectratio]{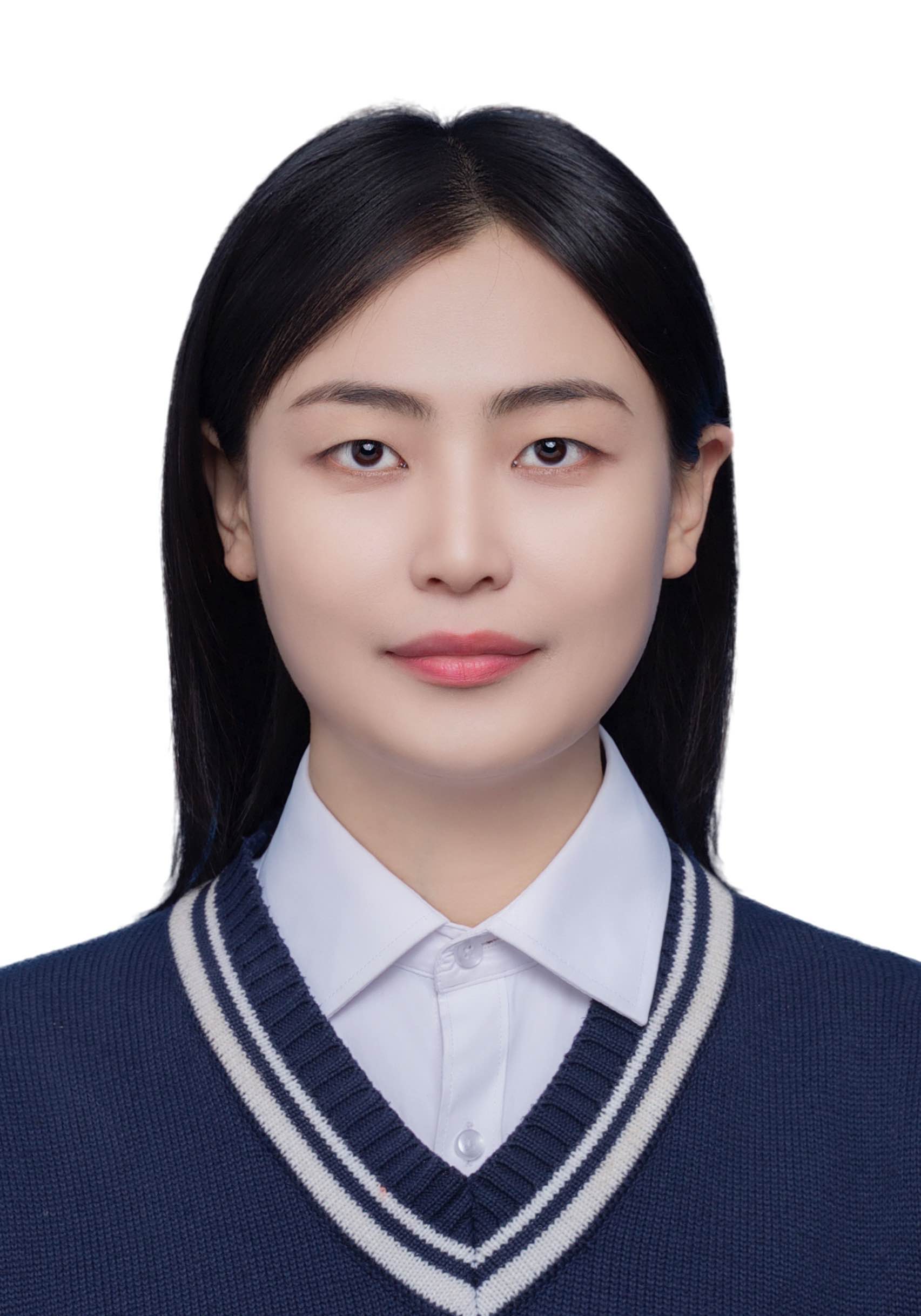}}]{Xin Zhang}
received the B.E. degree in Telecommunications Engineering from Xidian University, Xi'an, China in 2019. She is currently pursuing the Ph.D. degree with the Image Coding and Processing Center at State Key Laboratory of Integrated Services Networks, Xidian University, Xi'an, China. Her research interests span efficient deep learning, machine learning, and computer vision. Specifically, she is interested in model compression for computer vision models (CNN, ViT), knowledge distillation to both models and datasets, and general CV tasks (foundational model training and downstream applications).
\end{IEEEbiography}
\begin{IEEEbiography}[{\includegraphics[width=1in,height=2in,clip,keepaspectratio]{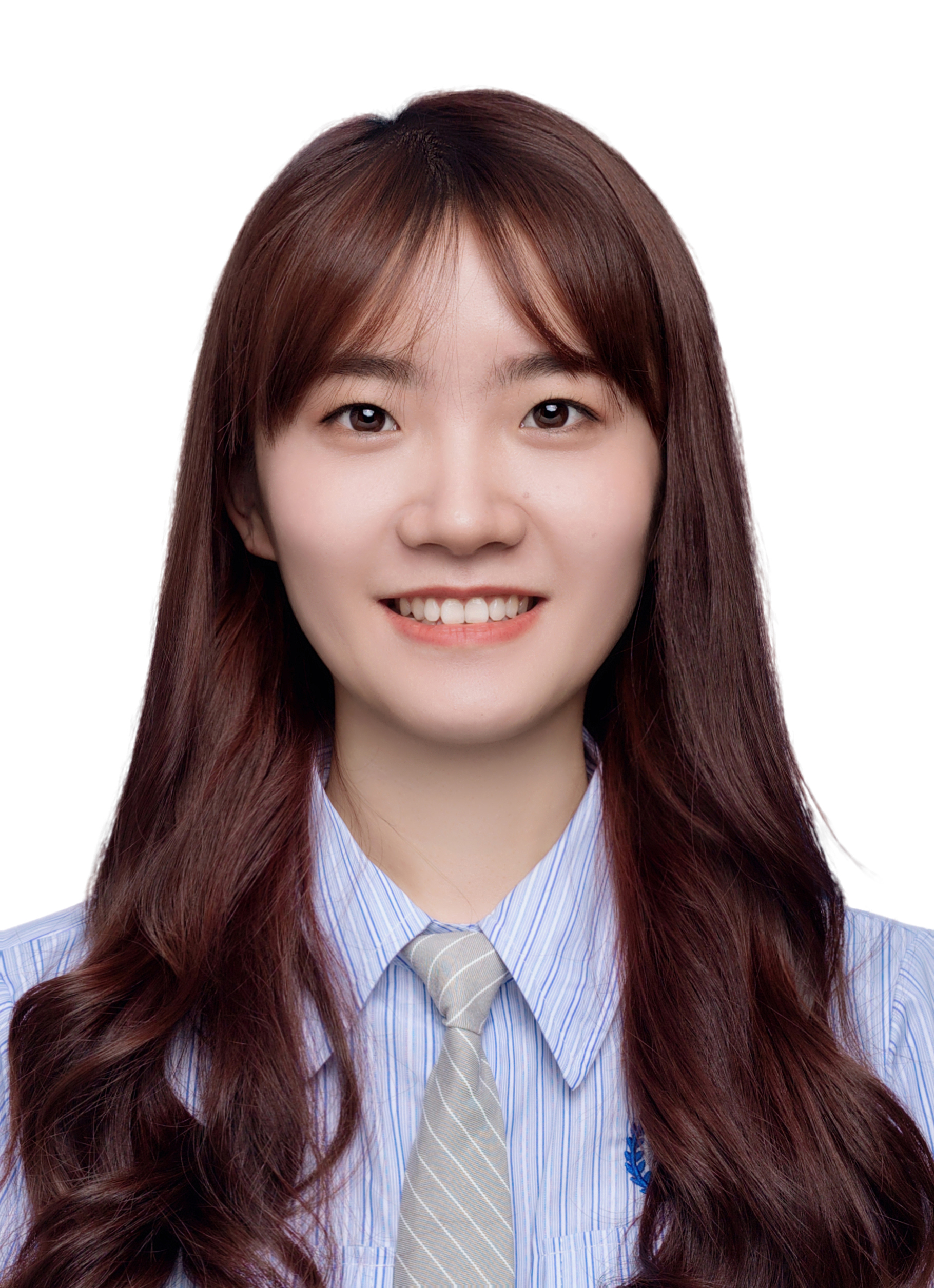}}]{Jiaqing Zhang}
received the B.E. degree in Telecommunications Engineering from Ningbo University, Zhejiang, China in 2019. She is currently pursuing the Ph.D. degree with the Image Coding and Processing Center at State Key Laboratory of Integrated Service Network, Xidian University, Xi'an, China.
Her research interests include multimodal image processing, remote sensing object detection, and network compression.
\end{IEEEbiography}
\begin{IEEEbiography}[{\includegraphics[width=1in,height=2in,clip,keepaspectratio]{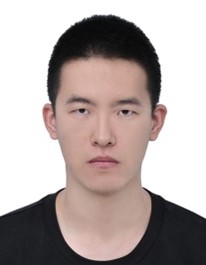}}]{Kai Jiang}
received the B.E. degree in information engineering and the Ph.D. degree in communication and information systems from Xidian University, Xi’an, China, in 2019 and 2024, respectively. His research interests include image processing and deep learning.
\end{IEEEbiography}
\vspace{-10.5cm}
\begin{IEEEbiography}[{\includegraphics[width=1in,height=2in,clip,keepaspectratio]{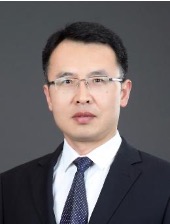}}]{Jie Lei}
received his M.S. degree in Telecommunications and Information Systems and his Ph.D. degree in Signal and Information Processing from Xidian University, China, in 2006 and 2010, respectively. He was a Visiting Scholar at the Department of Computer Science at the University of California, Los Angeles, USA, from 2014 to 2015. He served as a Professor at the School of Telecommunications Engineering, Xidian University, until 2023. Currently, he is a Research Fellow at the School of Electrical and Data Engineering at the University of Technology Sydney.
His research interests include wireless communication, remote sensing image processing, machine learning, and customized computing for big data applications.
\end{IEEEbiography}
\vspace{-10.5cm}
\begin{IEEEbiography}[{\includegraphics[width=1in,height=2in,clip,keepaspectratio]{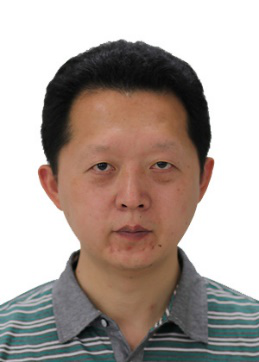}}]{Yunsong Li}
(Member, IEEE) received the M.S. degree in telecommunication and information systems and the Ph.D. degree in signal and information processing from Xidian University, China, in 1999 and 2002, respectively. He joined the School of Telecommunications Engineering, Xidian University in 1999 where he is currently a Professor. Prof. Li is the director of the image coding and processing center at the State Key Laboratory of Integrated Service Networks. His research interests focus on image and video processing and high-performance computing.
\end{IEEEbiography}

\end{document}